# Exploring Subnetwork Interactions in Heterogeneous Brain Network via Prior-Informed Graph Learning

Siyu Liu[1,2], Guangqi Wen[3,4], Peng Cao[1,2,5](✉), Jinzhu Yang[1,2,5], Xiaoli Liu[6], Fei Wang[7](✉), and Osmar R. Zaiane[8]

[1] Computer Science and Engineering, Northeastern University, Shenyang, China
[2] Key Laboratory of Intelligent Computing in Medical Image of Ministry of Education, Northeastern University, Shenyang, China
[3] School of Computer Science and Artificial Intelligence, Shandong Normal University, Jinan, China
[4] Shandong Key Laboratory of Intelligent Collaborative Perception and Computing, Jinan, China
[5] National Frontiers Science Center for Industrial Intelligence and Systems Optimization, Shenyang, China
caopeng@cse.neu.edu.cn
[6] AiShiWeiLai AI Research, Beijing, China
[7] Early Intervention Unit, Department of Psychiatry, Affiliated Nanjing Brain Hospital, Nanjing Medical University, Nanjing, China
fei.wang@yale.edu
[8] Amii, University of Alberta, Edmonton, AB, Canada

**Abstract.** Modeling the complex interactions among functional subnetworks is crucial for the diagnosis of mental disorders and the identification of functional pathways. However, learning the interactions of the underlying subnetworks remains a significant challenge for existing Transformer-based methods due to the limited number of training samples. To address these challenges, we propose **KD-Brain**, a Prior-Informed Graph Learning framework for explicitly encoding prior knowledge to guide the learning process. Specifically, we design a Semantic-Conditioned Interaction mechanism that injects semantic priors into the attention query, explicitly navigating the subnetwork interactions based on their functional identities. Furthermore, we introduce a Pathology-Consistent Constraint, which regularizes the model optimization by aligning the learned interaction distributions with clinical priors. Additionally, KD-Brain leads to state-of-the-art performance on a wide range of disorder diagnosis tasks and identifies interpretable biomarkers consistent with psychiatric pathophysiology. Our code is available at https://github.com/LiuSy0731/KDBrain.

# 1 Introduction

The identification of functional pathways, defined as distinct transmission circuits arising from the dynamic coupling of brain regions, is fundamental to psychiatric diagnosis [7,14,19]. Unlike static functional connectivity, these pathways describe the trajectory of information flow across the brain network, serving as robust biomarkers for distinguishing complex mental disorders, such as Autism Spectrum Disorder (ASD), Bipolar Disorder (BD), and Major Depressive Disorder (MDD).

To capture these functional pathways, we formulate the brain network as a heterogeneous graph and model the complex interactions among specialized functional subnetworks (e.g., DMN, CEN, and SN) via heterogeneous graph representation learning [5,18]. The diagnosis of mental disorders lies in precisely identifying critical disorder-related functional pathways in subnetwork interactions, for instance, MDD patients exhibit weakened "Default Mode Network (DMN)→Central Executive Network (CEN)→Salience Network (SN)" pathways linked to emotion regulation deficits [14]. In our work, we employ Transformers to learn the associations among subnetworks. However, under the constraints of limited medical samples, a purely data-driven attention mechanism lacks sufficient inductive bias and leads to poor generalization [9,16]. The reasons for this are: 1. **Unawareness of brain organization**, the positional encoding distinguishes the subnetworks using either generic one-hot encodings or physical 3D coordinates, which do not reflect the brain's functional organization. 2. **A large number of complex correlations**, the model may overfit to spurious correlations rather than learning the potential pathological mechanisms, resulting in subnetwork interaction patterns that lack neurobiological plausibility and generalization.

Both issues limit the model's ability to identify disease-related subnetwork interaction patterns that are essential for mental disorder diagnosis. To solve the above issues, we propose KD-Brain, a prior-informed graph learning framework designed to transform the interaction modeling from a blind correlation learning manner into a pathology-aware inference process guided by the pathophysiology of mental disorders. Specifically, KD-Brain extracts subnetwork-level topological features and learns their interactions by injecting dual priors: mapping semantic identities into attention queries and aligning interaction patterns with clinical constraints. Our contributions are summarized as follows:

- **New Perspective:** To address the challenge of limited medical data, we shift the paradigm of brain network analysis from "blind" statistical fitting to knowledge-guided learning, effectively modeling subnetwork interactions to uncover latent functional pathways.
- **Methodological Innovation:** A key innovation of our work lies in the seamless integration of prior knowledge into the Transformer learning paradigm by: (1) enriching the semantic representation of Query in cross-attention with pathology-aware priors, and (2) regularizing the distribution of attention scores to focus on clinically relevant subnetwork interaction.

– **SOTA Performance:** KD-Brain significantly outperforms 12 baselines on ASD, BD, and MDD diagnosis tasks. Further interpretation analysis further confirms that our model captures interpretable subnetwork interaction patterns consistent with neuroscientific consensus.

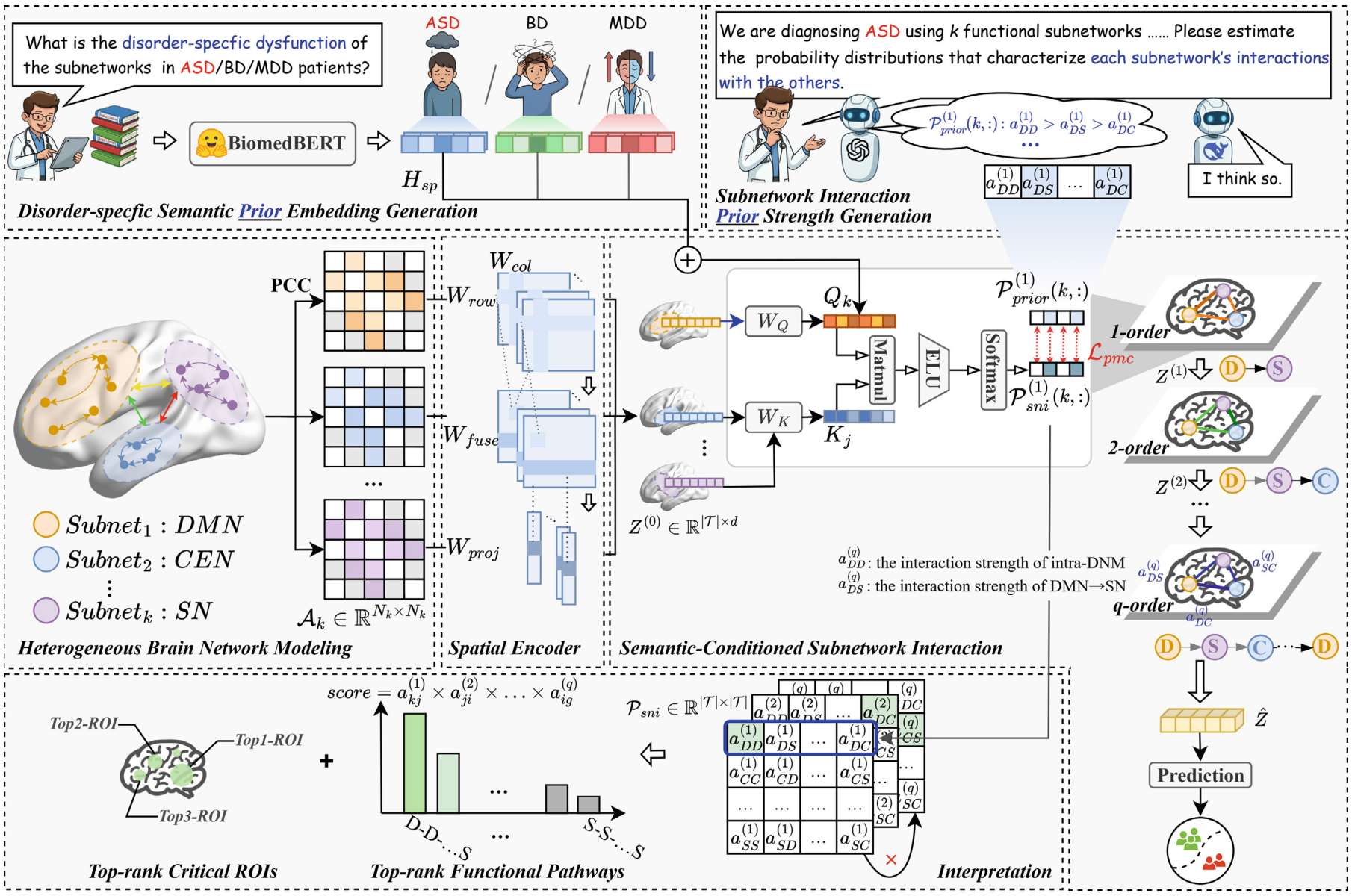

**Fig. 1.** Illustration of the KD-Brain.

## 2 Method

Figure 1. illustrates KD-Brain, a Prior-Informed Graph Learning framework for heterogeneous brain network analysis. It comprises three core components: (1) The Spatial Encoder for extracting intra-subnetwork topological embeddings; (2) Subnetwork Semantic Interaction Learning via disorder-specific semantic priors; (3) The Pathology-Consistent Constraint (PMC) to regularize the interaction modeling with prior clinical evidence for biological plausibility.

### 2.1 Preliminaries

We reformulate the brain network as a heterogeneous graph $\widetilde{\mathcal{G}} = \{\mathcal{G}_k\}_{k=1}^{|\mathcal{T}|}$, where $|\mathcal{T}|$ denotes the number of functional subnetworks. Each subnetwork $\mathcal{G}_k = \{\mathcal{V}_k, \mathcal{A}_k\}$ corresponds to a specific functional type, extracted via binary masks $\mathcal{M}_k \in \{0,1\}^N$. The subnetwork-specific connectivity is defined as $\mathcal{A}_k = \mathcal{M}_k \odot \mathcal{A} \in \mathbb{R}^{N_k \times N_k}$, where $N_k$ denotes the number of regions in $\mathcal{G}_k$.

### 2.2 Subnetwork Topology Representation Learning

To sufficiently capture the topology in the subnetworks, we leverage a bidirectional convolutional spatial encoder to extract complementary structural features at the subnetwork-level, which are then integrated into a refined embedding $Z_k \in \mathbb{R}^d$. Specifically, for the $k$-th subnetwork, its topological representation is obtained as follows:

$$\mathbf{H}_k = \sum_{c=1}^{C_{in}} \left( \mathbf{A}_{k,c} \mathbf{W}_{row,k,c} + \mathbf{A}_{k,c}^{\top} \mathbf{W}_{col,k,c} \right) \tag{1}$$

$$Z_k = (\text{LeakyReLU}\ (\mathbf{H}_k \mathbf{W}_{\text{fuse}}))\, \mathbf{W}_{\text{proj}} \tag{2}$$

where $W_{row,k,c}, W_{col,k,c} \in \mathbb{R}^{N_k \times C_{out}}$ denote the kernels for row and column convolutions, $W_{\text{fuse}} \in \mathbb{R}^{h \times C_{out}}$ represents the $1 \times 1$ convolution for channel compression, and $W_{\text{proj}} \in \mathbb{R}^{d \times h}$ is the projection matrix that maps the pooled representations into the latent embedding space.

### 2.3 Subnetwork Semantic Interaction Learning (SSIL)

**Disorder-Specfic Semantic Prior Embedding Generation.** Unlike previous methods that solely utilize the classification labels, we construct the disorder-specific semantic prior into the attention learning to explicitly guide the interactions among subnetworks. In our method, the description $S_k$ for the $k$-th subnetwork (e.g., DMN) is generated by involving the neurocognitive role and the disease-specific pathological alteration. For example,

(1) **In ASD diagnosis task:** The DMN is described as "*Responsible for social cognition... often shows hypoconnectivity leading to deficits in social interaction*".

(2) **In MDD diagnosis task:** The same DMN is redefined as "*Involved in self-referential processing...exhibits hyperconnectivity causing excessive negative self-focus and depressive rumination*".

These disorder-specfic descriptions are then encoded via BioMedBERT [2] to produce the semantic prior embedding: $H_{sp}^{(k)} = \text{BioMedBERT}(S_k) \in \mathbb{R}^{1\times d}$, and is subsequently aggregated into $H_{sp} = \left[ H_{sp}^{(1)}, \ldots, H_{sp}^{(|\mathcal{T}|)} \right]^{\top} \in \mathbb{R}^{|\mathcal{T}| \times d}$.

**Semantic-Conditioned Interaction.** This prior information, akin to conventional positional encoding, provides more semantically meaningful positional information within the brain network, i.e., functional signatures. Then, we reformulate the attention mechanism by integrating $H_{sp}^{(k)}$ directly into the Query vector at each interaction order ($l \in \{1, \ldots, q\}$) as:

$$Q_k^{(l)} = (Z_k^{(l-1)} + \lambda_{sp} \cdot H_{sp}^{(k)}) W_Q, \quad K_j^{(l)} = Z_j^{(l-1)} W_K \tag{3}$$

where $Z_k^{(l-1)}$ denotes the representation of the $k$-th subnetwork at the $(l-1)$-order interaction, with $Z^{(0)}$ denoting the initial representation, $\lambda_{sp}$ is a hyperparameter controlling the strength of semantic prior injection, and $W_Q, W_K \in \mathbb{R}^{d \times d}$ are learnable projection matrices.

Subsequently, we compute the normalized attention coefficient $\alpha_{k,j}^{(l)}$, which quantifies the $l$-order subnetwork interaction strength between the $k$-th source subnetwork and the $j$-th target subnetwork, $\alpha_{k,j}^{(l)} = \frac{\exp\left(\frac{Q_k^{(l)}(K_j^{(l)})^\top}{\sqrt{d}}\right)}{\sum_{t=1}^{|\mathcal{T}|} \exp\left(\frac{Q_k^{(l)}(K_t^{(l)})^\top}{\sqrt{d}}\right)}$. With the obtained normalized attention coefficients $\alpha^{(l)}$, the representation of the $l$-th subnetwork fused with its interaction semantics is updated as follows:

$$Z_k^{(l)} = \alpha_{k,k}^{(l)} \cdot Z_k^{(l-1)} + \sum_{j \in \mathcal{N}(k)} \alpha_{k,j}^{(l)} \cdot Z_j^{(l-1)} \tag{4}$$

Based on these coefficients, we construct the subnetwork interaction strength distribution at the $l$-th order: $\mathcal{P}_{sni}^{(l)}(k,:) = \left[\alpha_{k,1}^{(l)}, \alpha_{k,2}^{(l)}, \ldots, \alpha_{k,|\mathcal{T}|}^{(l)}\right] \in \mathbb{R}^{1 \times |\mathcal{T}|}$, which satisfies $\sum_{j=1}^{|\mathcal{T}|} \alpha_{k,j}^{(l)} = 1$ and quantifies the interaction strengths between the $k$-th subnetwork and all the other subnetworks. Finally, the representation of all subnetworks from the $q$-th order ( $q$ is a hyperparameter that controls the order number) denoted as $\hat{Z} = \left[Z_1^{(q)}, \ldots, Z_{|\mathcal{T}|}^{(q)}\right] \in \mathbb{R}^{|\mathcal{T}| \times d}$, is fed into an MLP classifier to produce the final predictions.

### 2.4 The Subnetwork Interaction Prior

To avoid the learned interactions that lack neurobiological plausibility, we incorporate clinical prior of subnetwork interaction as guidance. Specifically, we prompt an LLM (e.g., GPT-4) with a prompt $\mathcal{S}'$ involving the disorder description and the definitions of relevant functional subnetworks, yielding the prior interaction distribution for the $k$-th subnetwork: $\mathcal{P}_{prior}(k,:) = \text{LLM}(\mathcal{S}')^{(k)} \in \mathbb{R}^{1 \times |\mathcal{T}|}$, which represents a strength distribution over subnetworks derived from clinical experiences. Notably, we find that both GPT-4 and DeepSeek-R1 yield a consistent ranking of subnetwork interaction strengths.

To incorporate the subnetwork interaction prior during training, we introduce the Pathology-Consistent Constraint (PMC) based on Kullback–Leibler divergence, which aligns the learned subnetwork interaction patterns $\mathcal{P}_{sni}$ with the prior distribution $\mathcal{P}_{prior}$:

$$\mathcal{L}_{pmc} = \sum_{k=1}^{|\mathcal{T}|} \sum_{j=1}^{|\mathcal{T}|} \mathcal{P}_{prior}(k,j) \log \frac{\mathcal{P}_{prior}(k,j)}{\mathcal{P}_{sni}(k,j)} \tag{5}$$

Finally, the objective function is a weighted combination of the classification loss (Cross-Entropy) and the knowledge regularization loss:

$$\mathcal{L}_{total} = \mathcal{L}_{ce} + \beta \cdot \mathcal{L}_{pmc} \tag{6}$$

where $\beta$ is a hyperparameter balancing the contribution of data-driven learning and prior knowledge constraints.

**Table 1.** Performance comparison across three disorder diagnosis tasks, reported as mean ± std (%) for ACC and AUC; best results are highlighted in bold.

| Model | ABIDE (NYU) | NMU center | NMU center |
|---|---|---|---|
| | HC vs. ASD | HC vs. BD | HC vs. MDD |
| | ACC (AUC) | ACC (AUC) | ACC (AUC) |
| SVM-RFE [4] | 69.4±1.3 (68.4±0.9) | 70.9±1.1 (71.6±1.3) | 72.1±3.2 (72.0±1.8) |
| FNDL [3] | 66.9±1.4 (65.8±1.1) | 70.9±1.7 (70.4±1.4) | 71.1±0.5 (70.4±0.8) |
| MDGL [13] | 67.1±2.9 (64.1±3.2) | 69.8±2.7 (71.2±2.2) | 69.1±4.8 (70.8±3.7) |
| STAGIN [11] | 71.1±4.9 (63.2±3.7) | 72.4±1.2 (64.5±2.9) | 73.1±3.4 (66.0±4.3) |
| BrainGNN [12] | 64.4±5.9 (65.1±4.7) | 71.4±1.8 (69.9±1.7) | 68.5±2.6 (69.3±4.1) |
| MVS-GCN [20] | 71.4±4.2 (72.8±3.3) | 69.2±0.8 (66.1±1.3) | 71.3±2.1 (71.5±2.8) |
| HeBrainGNN [17] | 63.4±5.6 (62.1±5.1) | 71.9±0.8 (72.0±3.2) | 66.8±1.3 (64.3±0.6) |
| EAG-RS [8] | 64.9±2.4 (70.2±1.3) | 73.0±1.4 (74.6±1.2) | 66.7±1.4 (71.1±0.9) |
| BNT [9] | 71.9±1.4 (72.9±1.4) | 75.1±1.2 (73.2±1.0) | 71.4±1.3 (70.9±1.4) |
| Com-BrainTF [1] | 71.0±1.5 (71.6±1.5) | 74.0±1.4 (73.9±1.2) | 70.6±1.4 (72.4±1.5) |
| CAGT [15] | 73.2±1.1 (74.9±1.1) | 76.3±1.0 (75.1±0.9) | 72.7±1.0 (73.6±1.1) |
| KD-Brain ($q = 1$) | 77.3±1.9 (76.4±4.8) | 75.0±3.2 (75.3±2.2) | **73.3±0.4** (71.7±0.4) |
| KD-Brain ($q = 2$) | **78.7±1.5 (77.8±1.6)** | **76.8±2.8 (77.3±4.2)** | 70.25±1.5 (70.8±2.3) |
| KD-Brain ($q = 3$) | 76.8±0.5 (77.6±0.6) | 74.0±0.4 (73.0±0.6) | 71.7±3.4 **(73.1±4.3)** |
| KD-Brain ($SSIL \rightarrow GAT$) | 73.7±1.1 (73.4±1.4) | 73.0±2.1 (76.4±2.6) | 68.4±0.4 (71.3±1.1) |
| KD-Brain w/o $H_{sp}$ | 73.4±0.3 (71.8±0.5) | 72.7±2.7 (73.8±3.1) | 69.6±0.2 (71.1±0.3) |
| KD-Brain w/o $\mathcal{L}_{pmc}$ | 74.0±0.4 (73.4±0.6) | 74.0±3.2 (73.7±3.8) | 70.7±2.3 (71.0±4.8) |

# 3 Experiments and Interpretation

## 3.1 Dataset and Experimental Details

We evaluate our model on two datasets: the Autism Brain Imaging Data Exchange (ABIDE) and a single-center dataset from Nanjing Medical University (NMU center), covering 3 diagnosis tasks: ASD, BD, and MDD. The NMU center[1] dataset includes 246 healthy controls, 151 MDD patients, and 126 BD patients, all acquired at a single site under consistent inclusion and exclusion criteria. For ABIDE, to mitigate inter-site variability, we selected 172 subjects from the NYU site, comprising 74 ASD patients and 98 healthy controls. All fMRI data were preprocessed using DPABI [21], and the brain was parcellated into 116 regions based on the AAL atlas and further grouped into three functional subnetworks: the default mode network (DMN), salience network (SN), and central executive network (CEN).

[1] Approved by the Medical Science Ethics Committee of the Affiliated Brain Hospital of Nanjing Medical University (Ref. 2020-KY027-01).

### 3.2 Results and Discussion

**Comparison with SOTAs.** To evaluate the effectiveness of KD-Brain, we compared it against 12 state-of-the-art baseline methods, with the results summarized in Table 1. Our analysis reveals several key insights: (1) We observe that GNN-based models (e.g., HeBrainGNN, MVS-GCN) often achieve suboptimal performance, exemplified by HeBrainGNN dropping to 71.9%(ACC) on BD. They focused on local topological fitting within fixed structures neglects high-order interactions between functional subnetworks. (2) Transformer-based models (e.g., BNT, Com-BrainTF, CAGT) improve results by capturing global dependencies, with CAGT reaching 73.2% (ACC) on ASD. However, their unguided attention mechanisms often overfit to spurious correlations in small-sample datasets

**Ablation Analysis.** Table 1 validates 3 key findings: (1) We find that 2-order subnetwork interactions ($q = 2$) achieve the best overall performance in ASD and BD diagnosis tasks. (2) Our results indicate that replacing the proposed semantic interaction learning with a standard GAT leads to a substantial performance decline (e.g., 5.0%↓ on ASD), suggesting that unguided attention is insufficient to capture pathological heterogeneity. (3) We observe that removing either the semantic prior $H_{sp}$ or the pathological constraint $\mathcal{L}_{pmc}$ significantly degrades performance, highlighting the essential role of all components.

### 3.3 Effectiveness of Semantic-Conditioned Interaction

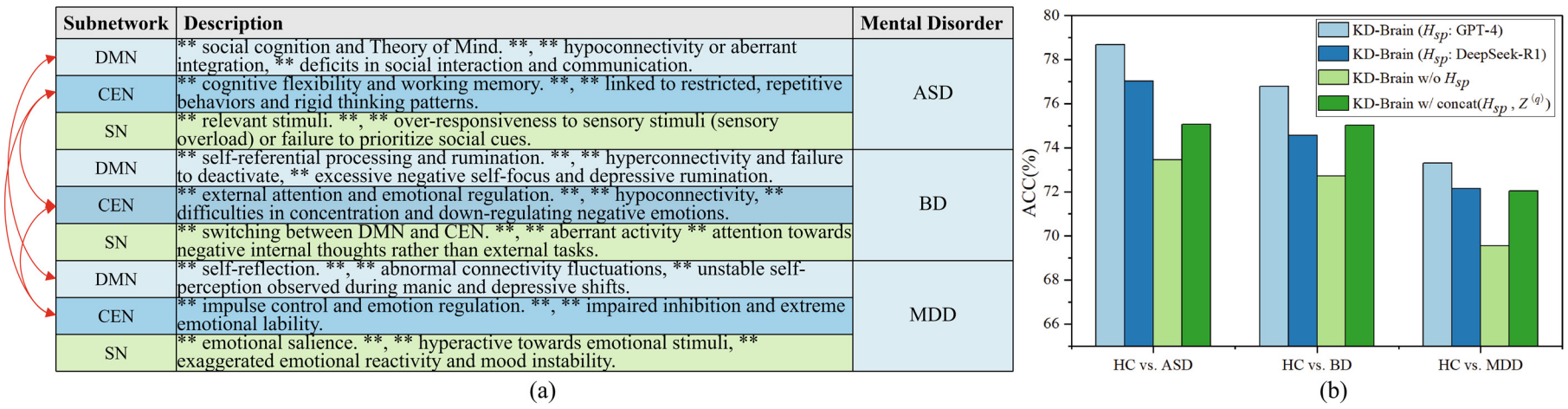

| Subnetwork | Description | Mental Disorder |
|---|---|---|
| DMN | ** social cognition and Theory of Mind. **, ** hypoconnectivity or aberrant integration, ** deficits in social interaction and communication. | ASD |
| CEN | ** cognitive flexibility and working memory. **, ** linked to restricted, repetitive behaviors and rigid thinking patterns. | |
| SN | ** relevant stimuli. **, ** over-responsiveness to sensory stimuli (sensory overload) or failure to prioritize social cues. | |
| DMN | ** self-referential processing and rumination. **, ** hyperconnectivity and failure to deactivate, ** excessive negative self-focus and depressive rumination. | BD |
| CEN | ** external attention and emotional regulation. **, ** hypoconnectivity, ** difficulties in concentration and down-regulating negative emotions. | |
| SN | ** switching between DMN and CEN. **, ** aberrant activity ** attention towards negative internal thoughts rather than external tasks. | |
| DMN | ** self-reflection. **, ** abnormal connectivity fluctuations, ** unstable self-perception observed during manic and depressive shifts. | MDD |
| CEN | ** impulse control and emotion regulation. **, ** impaired inhibition and extreme emotional lability. | |
| SN | ** emotional salience. **, ** hyperactive towards emotional stimuli, ** exaggerated emotional reactivity and mood instability. | |

**Fig. 2.** (a) The correlation between prior disorder-agnostic subnetwork embeddings, where red lines represent significant correlations ($P < 0.05$). (b) The effectiveness of Semantic-Conditioned Interaction. (Color figure online)

The disorder-specific description ("*" denotes disorder-agnostic term) is removed to generate the disorder-agnostic subnetwork prior embeddings. As shown in Fig. 2.(a), Pearson correlation analysis on these disorder-agnostic embeddings ($P < 0.05$) indicates that the statistical significance of correlation exists among prior embeddings of identical subnetworks across different disorders, demonstrating that LLM-generated priors accurately capture the intrinsic functional identity of subnetworks. Moreover, quantitative comparisons in Fig. 2.(b) clarify the distinct contributions of $H_{sp}$: (1) The benefits of semantic prior, regardless of GPT-4 or DeepSeek-R1 are witnessed, confirming that semantic priors are essential for capturing complex pathological patterns beyond purely data-driven features. (2) The necessity of injecting semantic prior into the attention Query acts as an enhanced "navigator" for subnetwork interactions, compared with the simple concatenation (concat($H_{sp}, Z^{(q)}$)).

(a)

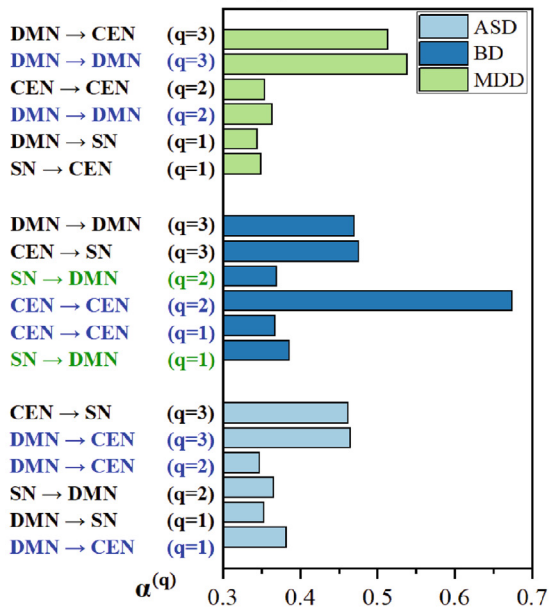

(b)

**Critical Brain Region**

| ASD | BD | MDD |
|---|---|---|

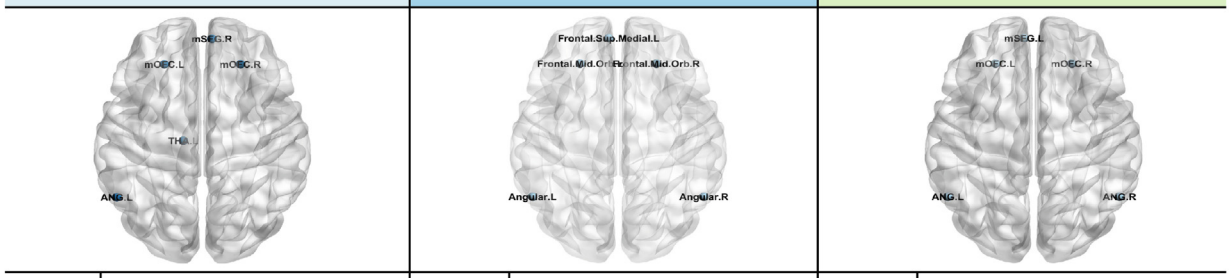

| Rank | Region Name | Rank | Region Name | Rank | Region Name |
|---|---|---|---|---|---|
| 1 | Angular_R | 1 | Thalamus_L | 1 | Precuneus_L |
| 2 | Angular_L | 2 | Frontal_Mid_Orb_R | 2 | Frontal_Mid_Orb_R |
| 3 | Frontal_Mid_Orb_R | 3 | Frontal_Mid_Orb_L | 3 | Frontal_Sup_Orb_L |
| 4 | Frontal_Mid_Orb_L | 4 | Frontal_Sup_Medial_R | 4 | Frontal_Mid_Orb_L |
| 5 | Frontal_Sup_Medial_L | 5 | Angular_L | 5 | Thalamus_L |

**Critical Functional Pathway**

| | q = 1 | | q = 2 | | q = 3 | |
|---|---|---|---|---|---|---|
| | Functional Pathway | Score | Functional Pathway | Score | Functional Pathway | Score |
| ASD, Top-1 | DMN→CEN | 0.3816 | DMN→CEN→DMN | 0.1296 | DMN→CEN→DMN→CEN | 0.0603 |
| ASD, Top-2 | DMN→SN | 0.3531 | DMN→SN→DMN | 0.1292 | DMN→SN→DMN→CEN | 0.0601 |
| BD, Top-1 | SN→DMN | 0.3859 | CEN→CEN→CEN | 0.2481 | CEN→CEN→CEN→SN | 0.1180 |
| BD, Top-2 | CEN→CEN | 0.3678 | SN→CEN→CEN | 0.2211 | SN→CEN→CEN→SN | 0.1051 |
| MDD, Top-1 | SN→CEN | 0.3494 | SN→CEN→CEN | 0.1236 | DMN→DMN→DMN→DMN | 0.0665 |
| MDD, Top-2 | DMN→SN | 0.3442 | SN→CEN→SN | 0.1192 | DMN→DMN→DMN→CEN | 0.0634 |

(c)

**Fig. 3.** The visualization of multi-level biomarkers identified by KD-Brain. (a) Top-2 critical subnetwork interaction strengths, quantified by the normalized attention coefficients $\alpha_{k,j}^{(q)}$ in $\mathcal{P}_{sni}^{(q)}$. (b) Top-5 critical brain regions. (c) Top-2 functional pathways, where the pathway score is calculated as the joint probability of sequential interactions across orders: $Score = \prod_{l=1}^{q} \alpha_{k,j}^{(l)}$, where $q = 1, 2, 3$, respectively.

### 3.4 Interpretation

**Macroscopic Level: Subnetwork Interactions.** Figure. 3.(a) demonstrates that our model captures distinct subnetwork interactions in different orders ($q = 1/2/3$), Fig. 3.(c) further maps these into functional pathways. For ASD, the identified functional pathway "DMN→CEN→DMN" reflects the inability of the patient to disengage from internal processing, which in turn interferes with the executive network responsible for external attention (CEN), revealing the pathological mechanism of the blockage of external social cognition due to the dominance of internal thinking [10]. For BD, "CEN→CEN→CEN" implies pathological rigidity within the executive system that reflects the cognitive disruption that occurs during mood episodes. This circuit abnormality is closely related to impaired working memory, planning, and rule-based learning [22]. For MDD, SN-initiated pathways (e.g., SN→CEN→CEN) reflect the pathological characteristics of MDD patients with decreased control efficiency while performing cognitive tasks [14,18].

**Microscopic Level: Critical Brain Region.** Figure. 3.(b) illustrates that the Thalamus and Orbitofrontal Cortex are highlighted as top-ranked biomarkers for both BD and MDD. This overlap provides computational evidence for the shared neuropathology of mental disorders [6,22]. These findings align closely with established neurobiological literature, thereby reinforcing the validity and biological plausibility of our conclusions.

## 4 Conclusion

In this paper, we proposed **KD-Brain**, a prior-informed graph learning framework. By synergizing the Semantic-Conditioned Interaction mechanism, which navigates subnetwork interactions via semantic priors, and the Pathology-Consistent Constraint, which aligns learned interaction distributions with clinical priors, our framework captures biologically plausible subnetwork interaction patterns while preventing overfitting in limited data. Extensive evaluations on ASD, BD, and MDD diagnosis tasks demonstrate that KD-Brain achieves state-of-the-art performance, while identifying interpretable functional pathways and critical brain regions that align closely with psychiatric pathophysiology.

**Acknowledgments.** This research was supported by the National Natural Science-Foundation of China (62076059) and the Science and Technology Joint Project of Liaoning Province (2023JH2/101700367 and ZX20240193).

**Disclosure of Interests.** The authors have no competing interests to declare that are relevant to the content of this article.